# Optimization of bi-directional gated loop cell based on multi-head attention mechanism for SSD health state classification model

Zhizhao Wen*
*Department of Computer Science*
Rice University
Houston, United States
zhizhaowen@alumni.rice.edu*

Ruoxin Zhang
*Department of Computer Science*
Rice University
Houston, United States
lz37@alumni.rice.edu

Chao Wang
*Department of Computer Science*
Rice University
Houston, United States
zylj2020@outlook.com

***Abstract*—Aiming at the critical role of SSD health state prediction in data reliability assurance, this study proposes a bidirectional gated recurrent unit (BiGRU-MHA) hybrid model incorporating a multi-head attention mechanism, which effectively enhances the accuracy and stability of storage device health classification prediction by innovatively integrating temporal feature extraction and key information focusing capabilities. The model utilizes the bidirectional timing modeling advantage of BiGRU network to capture the forward and backward dependencies of SSD degradation features, and at the same time introduces the multi-head attention mechanism to dynamically assign feature weights to enhance the identification of sensitive indicators of health status. The experimental results show that the proposed model achieves 92.70% and 92.44% classification accuracy on the training set and test set, respectively, with a difference of only 0.26%, demonstrating excellent model generalization performance. Further analyzed by the subject work characteristic curve (ROC), the area under the curve (AUC) on the test set reaches 0.94, which confirms that the model has a highly robust binary classification discriminative ability. This study not only provides a new technical path for SSD health prediction but also breaks through the bottleneck of the traditional model in terms of the performance difference between the training-testing set with a generalization error of only 0.26%, which is of great practical value for the preventive maintenance of industrial-grade storage systems. The result can significantly reduce the probability of data loss by warning potential failure risks in advance, while optimizing the maintenance cost, providing verifiable intelligent decision support for building a highly reliable computer storage system, which is widely applicable to the health management of cloud computing data centers and edge storage devices.***

## I. INTRODUCTION

With the acceleration of the digitization process, the demand for data storage is growing exponentially, and the reliability of storage media has become a core issue in ensuring data security [1]. Traditional storage devices (e.g., mechanical hard disk HDD, solid-state hard disk SSD) gradually deteriorate in long-term operation due to physical wear and tear, aging of electronic components, environmental interference, and other factors, resulting in the risk of data loss or system crash. Especially in data center, cloud computing and edge computing scenarios, the large-scale deployment of storage devices makes manual monitoring of health status inefficient and costly [2]. Earlier studies relied on threshold alarms or statistical models to predict failures, but these methods are not accurate and timely enough under complex operating conditions to cope with the diverse failure modes of new storage media [3]. Therefore, there is an urgent need for an intelligent method that can dynamically assess the health state of storage media and achieve accurate classification to support preventive maintenance and resource optimization.

Machine learning provides an efficient technical path for health degree classification by mining the hidden laws in storage media operation data. First, supervised learning-based algorithms are able to learn the mapping relationship between features and health states from historical failure data, and build high-precision classification models by combining multivariate data such as SMART parameters, I/O performance logs, and temperature [4]. Second, deep learning excels in processing time-series signals and complex nonlinear relationships, capturing the dynamic trend of storage media performance degradation and realizing early anomaly detection, particularly when using adaptive training strategies to enhance model adaptability and generalization [5][6]. In addition, unsupervised learning is capable of identifying potential failure modes without relying on labeled data, which is suitable for scenarios lacking a priori knowledge. Through feature engineering and model optimization, machine learning not only improves the real-time performance of healthiness classification, but also reduces the false alarm rate and provides decision support for proactive operation and maintenance of storage systems.

Although machine learning shows potential in storage healthiness classification, it still faces challenges such as data quality and model generalizability. With the iteration of storage media technologies (e.g., SCM, DNA storage), machine learning algorithms need to evolve continuously to cope with more complex health assessment needs, and ultimately drive the storage system towards highly reliable and adaptive development. In this paper, we optimize the bidirectional gated recurrent unit model based on the multi-head attention mechanism for classifying and predicting the health status of SSDs. SSD health prediction can effectively prevent data loss, avoid system crashes, and optimize maintenance timing by warning potential failures in advance, thus significantly improving the reliability, stability, and storage management efficiency of computer systems.

## II. DATASET SOURCES AND SELECTED DATASETS

In this paper, we conduct experiments using a private dataset that contains 593 SSD health status monitoring data with eight features and one triple categorical target variable (normal/warning/failure), in which the number of bad blocks reflects the degree of physical damage, segmented uniform percentage of remaining life (0-100%) serves as the core degradation indicator, and the bimodal temperature distribution indicates the normal versus abnormal cooling scenarios in combination with the wear parameters such as the amount of write and the number of erasures. The target variable is the health status of the SSD (normal, warning, failure), and the dataset can be used to develop failure prediction models, validate feature engineering strategies, and study unbalanced classification problems. Part of the dataset is shown in Table 1.

SELECTED DATA SETS.

| Usage duration | Average erase times | Total write volume | T | Read-write error rate | Power-on times | Classification of health status |
|---|---|---|---|---|---|---|
| 12060 | 14 | 37.7 | 37.6 | 0.11 | 3161 | normalcy |
| 9337 | 16 | 172.98 | 52.8 | 0.01 | 205 | Malfunction |
| 23276 | 16 | 41.6 | 49.3 | 0.03 | 4713 | Normal. |
| 18174 | 15 | 79.46 | 39.1 | 0.08 | 2039 | Normal. |
| 17190 | 9 | 143.42 | 66.1 | 0.07 | 4509 | Malfunction. |
| 8670 | 12 | 99.42 | 47.5 | 0.21 | 2925 | Normal |
| 13476 | 13 | 140.7 | 42.5 | 0.17 | 501 | Normal. |
| 16682 | 10 | 138.45 | 38.3 | 0.25 | 2950 | Normal |
| 16978 | 15 | 115.32 | 43.5 | 0.18 | 444 | Normal |
| 20532 | 14 | 57.68 | 44.4 | 0.21 | 672 | Early Warning |
| 17366 | 12 | 39.88 | 43.8 | 0.06 | 1209 | Normal |
| 27585 | 16 | 141.1 | 42.7 | 0.18 | 2936 | Normal |
| 6639 | 13 | 22.75 | 58.4 | 0.01 | 2947 | Failure |
| 16472 | 9 | 81.52 | 68.3 | 0.00 | 1309 | Early warning |

## III. METHOD

### A. Multihead Attention Mechanism

Multi-attention mechanism is a core component of the Transformer model, which aims to enhance the model's ability to capture complex patterns by parallelizing the processing of information associations at different semantic levels [7]. The core idea is to map the embedding vectors of the input sequences into different "subspaces" through multiple independent linear transformations. Each subspace calculates the attention weights and generates the corresponding contextual representations, and finally integrates all the results into a unified output through linear projection after stitching [8]. This process enables the model to simultaneously focus on the relevance of different locations and semantic dimensions in the sequence by decomposing a single attention into multiple independent perspectives. The network structure of the multiple attention mechanism is shown in Figure 1.

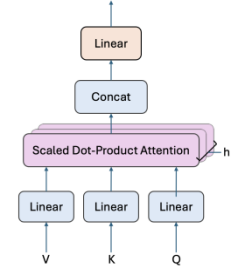

Fig. 1. The network structure of the multiple attention mechanism.

For specific implementation, the input vectors are first transformed by multiple linear transformations to generate multiple query, key, and value matrices, with each group corresponding to an attention head. Each head independently performs the scaled dot product attention computation: the attention weight matrix is generated by the similarity between the query and the key, and the value vectors are weighted and summed by Softmax normalization to obtain the output of that head. Subsequently, the outputs of all heads are spliced in the feature dimension and compressed back to the original dimension by a learnable projection matrix to maintain computational efficiency [9]. For example, if the input dimension is 512 and 8 heads are used, the output dimension of each head is 64, which is restored to 512 dimensions after splicing. This design ensures the flexibility of multi-view modeling and reduces the computational complexity through dimensional decomposition.

While the multi-head attention mechanism inherently lacks positional information and thus typically requires explicit positional encoding, the use of BiGRU mentioned below in our hybrid model addresses this limitation by naturally encoding sequential dependencies. BiGRUs capture temporal context through sequential state transitions, providing an implicit positional representation beneficial for modeling SSD degradation patterns.

### B. Bidirectional Gated Loop Cell

Bidirectional gated recurrent unit (BiGRU) is an improved model based on recurrent neural network (RNN), which can capture long-range dependencies in sequence data more efficiently by combining bi-directional information flow and gating mechanism [10]. Its core design idea lies in utilizing both forward and backward information of sequences and combining the gating unit to dynamically regulate the information flow, to improve the model's ability to interpret the context. The network structure of BiGRU is shown in Fig. 2.

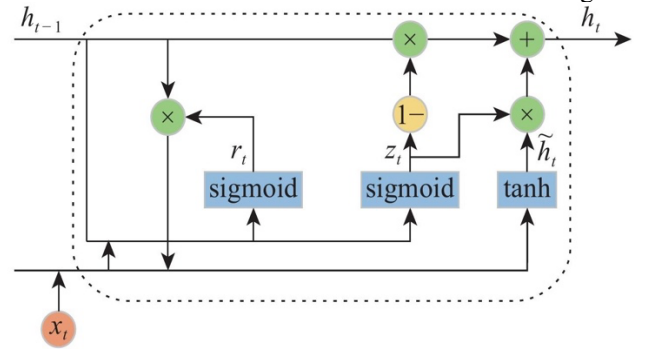

Fig. 2. The network structure of BiGRU.

Structurally, BiGRU consists of two independent GRU networks: one processes the input sequences in chronological order (forward GRU), and the other processes the input sequences in reverse order (reverse GRU).The GRU unit itself realizes the dynamic control of the information through two key structures, namely, the "update gate" and the "reset gate" [11]. The update gate determines how much of the historical memory is retained at the current moment, while the reset gate controls how much of the historical state is rewritten by the current input. This design allows GRU to avoid the gradient vanishing problem of traditional RNNs while being more concise than the LSTM structure. In the bi-directional structure, the GRUs in each of the two directions process the sequence information independently, and ultimately splice or weighted fuse the forward and backward hidden states to form a comprehensive representation containing complete contextual information.

### C. Optimization Mechanism of Multihead Attention Mechanism for BiGRUs

The multi-head attention mechanism captures long-distance dependencies and diverse semantic features in sequences from different subspaces by computing multiple independent attention heads in parallel, and when combining it with the bidirectional gated recurrent unit (BiGRU), firstly, the forward and reverse hidden states of BiGRU are used to model the local timing dependencies of the sequences respectively, and then the timing features outputted by BiGRU are globally cross-stepped through multi-head attention interaction and dynamic weight allocation, so as to strengthen the key position information and suppress the noise interference, and at the same time, overcome the long-distance correlation attenuation problem caused by the fixed window limitation of the traditional BiGRU, and ultimately realize the synergistic optimization of the local dynamic gating and the global context perception, so as to enhance the model's ability of modeling the complex sequence patterns.

The multi-head attention mechanism optimizes the context modeling capability of BiGRUs by capturing the semantic correlations of different subspaces in the sequence in parallel, and its core principle lies in combining the temporal feature extraction of bi-directional gated recurrent units (BiGRUs) with the global dependency modeling of the attention mechanism. In the structural design, the number of heads of multi-head attention is set to 3, which not only ensures that the model can focus on multiple types of interaction patterns between words, but also avoids the risk of parameter redundancy and overfitting due to the excessive number of heads, whereas the dimension of the hidden layer of the BiGRU needs to be coordinately configured with that of attention, which is set to be an integer multiple of h. 6, which is used to generate the matrix of query, key, and value through linear projection, realizing the multi-view decoupling of the feature space. The fusion of the two adopts the "feature-enhanced tandem architecture": first, BiGRU scans the input sequence in both directions, and generates the hidden state matrix containing the preceding and following text information; then H is inputted into the multi-head attention module for the cross-position weight calculation, and the attention outputs of each head are obtained by scaling the dot product of the attention; then all the attention outputs are spliced together and fused into the attention-enhanced features by the linear layer; finally, the original hidden state is fused with the residual connection to realize the multi-view decoupling. The final optimized context representation is obtained by summing the original hidden state H with the attention feature A through residual concatenation, followed by layer normalization. This design retains the advantages of BiGRU temporal modeling while using multi-head attention to break through the limitation of the local perceptual field of recurrent neural networks and explicitly establishes long-range dependencies. Especially when dealing with the complex interactions within sequences, the multi-attention heads can capture the differentiated features such as location-sensitive, content-related, etc., respectively, and the residual structure effectively mitigates the gradient vanishing and improves the training stability. The fusion model structure is shown in Fig. 3.

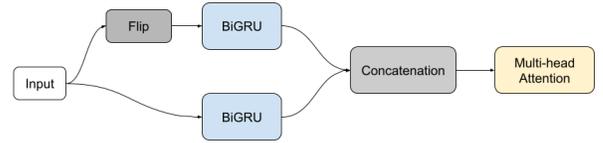

Fig. 3. The network structure of BiGRU.

### IV. RESULT

In this paper, we optimize BiGRU's classification model for SSD health status prediction using the multi-head attention mechanism, and evaluate the effectiveness of the model using the parameters of collinearity, precision, recall and F1. In terms of experimental parameter settings, this paper uses the Adam optimizer with the maximum training epoch set to 500, gradient threshold set to 1, initial learning rate set to 0.001, regularization parameter set to 0.001, and training using CPU. In software, Pytorch 1.13.1 was used and in hardware parameters, Apple M1 CPU with 8 cores and 16GB memory were used.

Firstly output the variation curves of the model's ACCURACY and LOSS during the training process, as shown in Fig. 4.

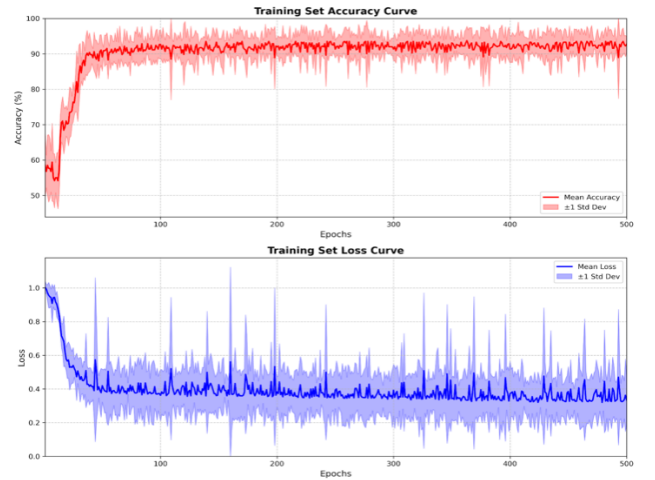

Fig. 4. The variation curves of the model's ACCURACY and LOSS.

Output the confusion matrix of the optimized BiGRU model with the multi-head attention mechanism proposed in this paper on the training set and the test set. The confusion matrix of the training set is shown in Fig. 5 and the confusion matrix of the test set is shown in Fig. 6.

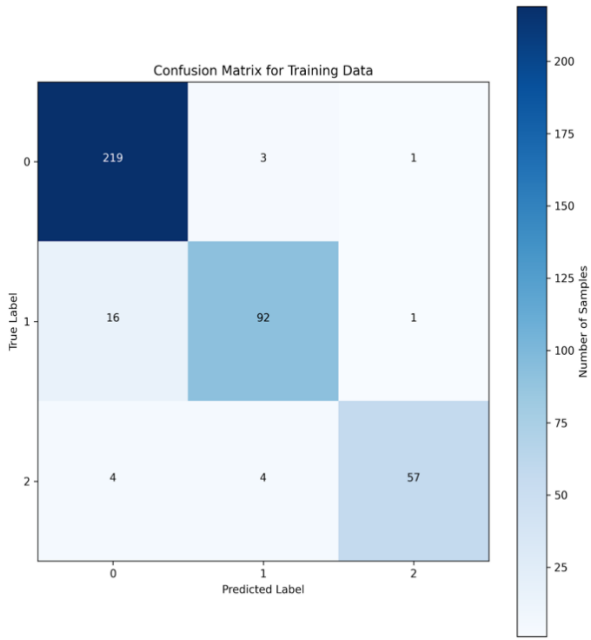

Fig. 5. The confusion matrix of the training set.

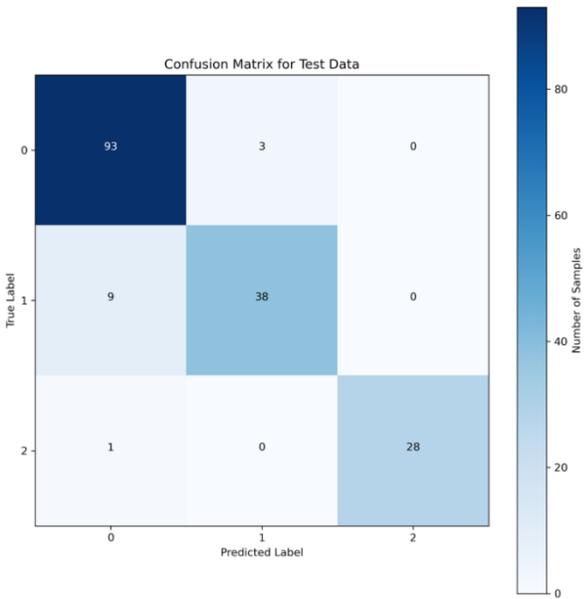

Fig. 6. The confusion matrix of the test set.

From the results of the confusion matrix, it can be seen that the accuracy of the model proposed in this paper is 92.70% on the training set and 92.44% on the test set, and the model proposed in this paper is able to predict the health state of the SSD very well, and the difference in the accuracy between the test set and the training set is 0.26%, which indicates that the model has a very good generalization ability.

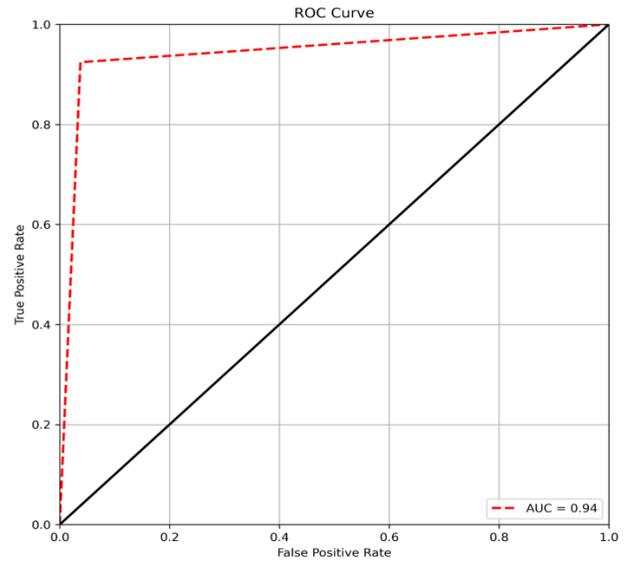

Fig. 7. The ROC curve of the output model,.

The ROC curve of the output model, as shown in Fig. 7, shows that the AUC value of the model is 0.94 from the ROC curve of the test set, indicating that the model proposed in this paper is able to predict the health state of the SSD well.

## V. CONCLUSION

In this study, a hybrid model of bidirectional gated recurrent unit (BiGRU-MHA) optimized based on multi-head attention mechanism is proposed for SSD health state prediction. By deeply integrating the multi-head attention mechanism with the bi-directional GRU network, the model not only captures the temporal features and long-term dependencies in the SSD operation data, but also adaptively focuses on key health indicators. The empirical results show that the constructed model achieves 92.70% and 92.44% prediction accuracies on the training set and the test set, respectively, with only a minor difference of 0.26% between the two, which fully verifies the superiority of the model in terms of feature extraction and generalization performance. Particularly noteworthy is that the high AUC value of up to 0.94 on the test set further corroborates the high differentiation of the model in health state classification prediction, which provides a new technological path for constructing a reliable health management system for SSDs.

The breakthroughs in this study have significant value for improving the reliability of storage systems. The prediction framework not only optimizes the data center operation and maintenance strategy and reduces the maintenance cost, but more importantly lays the algorithmic foundation for constructing an intelligent storage system with autonomous decision-making. Future research will focus on exploring the lightweight deployment of the model in edge computing environments, as well as the cross-device migration learning capability of multiple hard disk models.

## VI. CODE AVAILABILITY

We release our code and data to facilitate reproducibility at https://github.com/WindsorWZZ/BiGRU-multi-head-attention.

## VII. FUTURE WORK

Future research will focus on the following directions: first, the model architecture will be further optimized to explore the multimodal fusion of different attention mechanisms (sparse attention or hierarchical attention) and recurrent neural networks to improve the ability to capture SSD temporal degradation features; second, it is planned to build a larger multi-source heterogeneous dataset covering SSD operating data of different brands, capacities and usage scenarios; meanwhile, the model interpretability study will reveal key indicators of health state prediction to provide theoretical support for the study of SSD failure mechanisms; in addition, the model interpretability study will be conducted to use feature importance analysis to reveal key indicators of health state prediction. HDD operation data, and enhance the generalization performance of the model through migration learning; at the same time, model interpretability research will be carried out, and feature importance analysis will be used to reveal the key indicators of health state prediction, providing theoretical support for the study of SSD failure mechanisms; in addition, an online incremental learning framework is proposed to be developed to realize the dynamic updating of the model during the device operation process, and to enhance real-time prediction capability; finally, the technology will be explored Finally, we will explore the lightweight deployment scheme on edge computing devices, combine the knowledge distillation and other methods to balance the prediction accuracy and computational efficiency, and promote the engineering application of the research results in the actual operation and maintenance system.